\def\year{2021}\relax
\documentclass[letterpaper]{article} 
\usepackage{aaai21}  
\usepackage{times}  
\usepackage{helvet} 
\usepackage{courier}  
\usepackage[hyphens]{url}  
\usepackage{graphicx} 
\usepackage{graphicx}  
\usepackage{natbib}  
\usepackage{caption} 
\usepackage{amsmath}
\usepackage{amssymb}
\usepackage{multirow}
\usepackage{enumitem}
\usepackage[ruled]{algorithm2e}

\usepackage[switch]{lineno}

\title{Contrastive Transformation for Self-supervised Correspondence Learning}
\author {
    Ning Wang\textsuperscript{\rm 1},~
    Wengang Zhou\textsuperscript{\rm 1,2},~
    Houqiang Li\textsuperscript{\rm 1,2} \\
}
\affiliations {
    \textsuperscript{\rm 1} CAS Key Laboratory of GIPAS, University of Science and Technology of China \\
    \textsuperscript{\rm 2} Institute of Artificial Intelligence, Hefei Comprehensive National Science Center \\
    wn6149@mail.ustc.edu.cn, \{zhwg, lihq\}@ustc.edu.cn
}

\begin{document}

\maketitle


\begin{abstract}
	In this paper, we focus on the self-supervised learning of visual correspondence using unlabeled videos in the wild.
	Our method simultaneously considers intra- and inter-video representation associations for reliable correspondence estimation.
	The intra-video learning transforms the image contents across frames within a single video via the frame pair-wise affinity.
	To obtain the discriminative representation for instance-level separation, we go beyond the intra-video analysis and construct the inter-video affinity to facilitate the contrastive transformation across different videos.
	By forcing the transformation consistency between intra- and inter-video levels, the fine-grained correspondence associations are well preserved and the instance-level feature discrimination is effectively reinforced.
	Our simple framework outperforms the recent self-supervised correspondence methods on a range of visual tasks including video object tracking (VOT), video object segmentation (VOS), pose keypoint tracking, \emph{etc}. 
	It is worth mentioning that our method also surpasses the fully-supervised affinity representation (\emph{e.g.}, ResNet) and performs competitively against the recent fully-supervised algorithms designed for the specific tasks (\emph{e.g.}, VOT and VOS).
	
\end{abstract}

\begin{figure}[t]
	\centering
	\includegraphics[width=7.9cm]{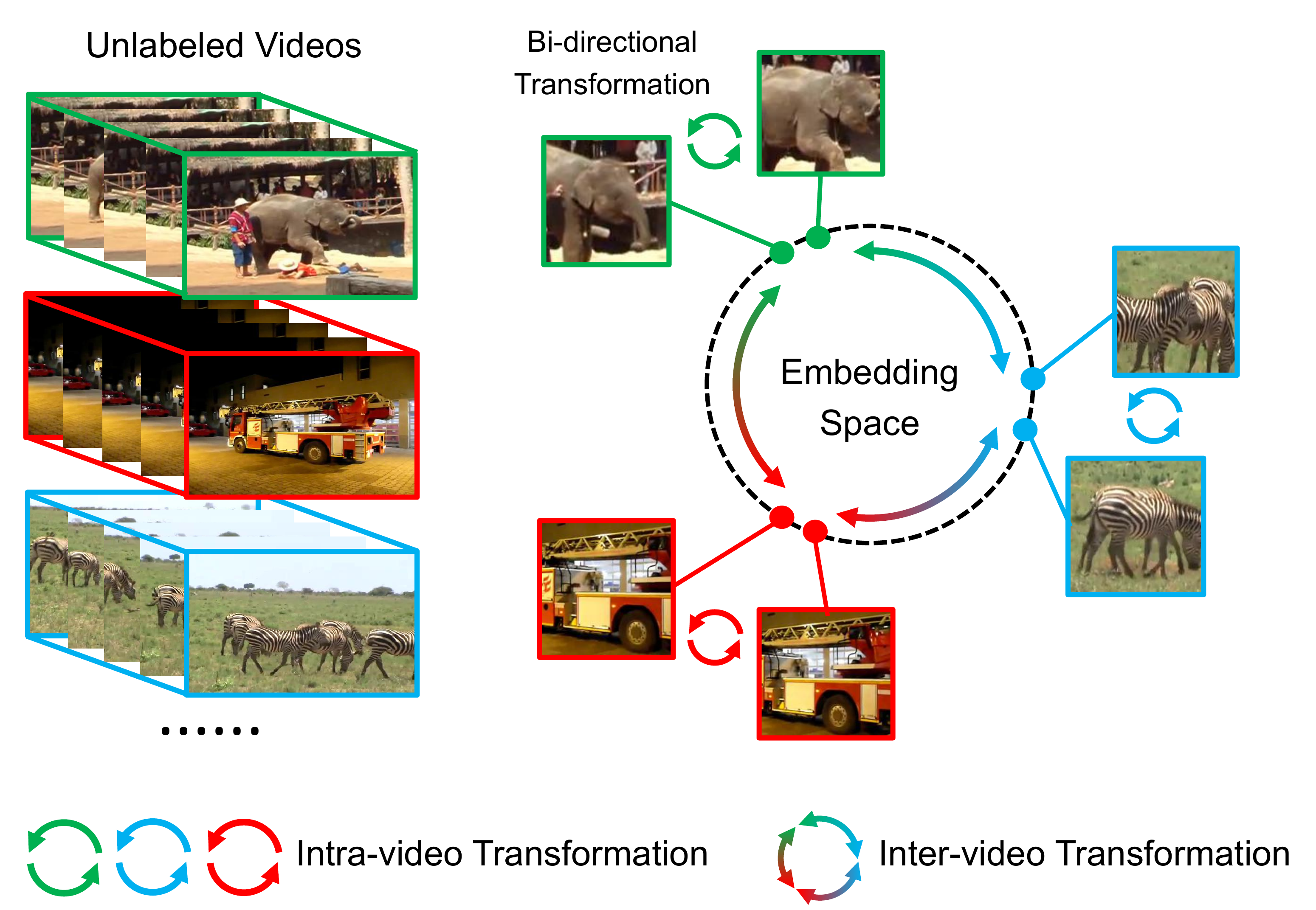}
	\caption{The proposed approach targets at learning correspondence using unlabeled videos. Previous works mainly focus on the content transformation within each video clip. Our framework simultaneously tracks (intra-video level) and spreads (inter-video level) the feature embeddings to preserve the fine-grained matching capability while encouraging the contrastive embedding learning.}\label{fig:1} 
\end{figure}

\section{Introduction}

Learning representations for visual correspondence is a long-standing problem in computer vision, which is closely related to many vision tasks including video object tracking, keypoint tracking, and optical flow estimation, \emph{etc}. 
This task is challenging due to the factors such as viewpoint change, distractors, and background clutter.

Correspondence estimation generally requires human annotations for model training.
Collecting dense annotations, especially for large-scale datasets, requires costly human efforts.
%
To leverage the large volume of raw videos in the wild, the recent advances focus on self-supervised correspondence learning by exploring the inherent relationships within the unlabeled videos.
In \cite{TimeCycle}, the temporal cycle-consistency is utilized to self-supervise the feature representation learning. 
To be specific, the correct patch-level or pixel-wise associations between two successive frames should match bi-directionally in both forward and backward tracking trajectories. 
The bi-directional matching is realized via a frame-level affinity matrix, which represents the pixel pair-wise similarity between two frames.
In \cite{colorizition,UVC}, this affinity is also utilized to achieve the content transformation between two frames for self-supervision. 
A straightforward transformation within videos is the color/RGB information.
More specifically, 
the pixel colors in a target frame can be ``copied'' (or transformed) from the pixels in a reference frame.
By minimizing the differences between the transformed and the true colors of the target frame, the backbone network is forced to learn robust feature embeddings for identifying correspondence across frames in a self-supervised manner.

In spite of the impressive performance, existing unsupervised correspondence algorithms put all the emphasis on the intra-video analysis.
Since the scenario in one video is generally stable and changeless, establishing the correspondence within the same videos is less challenging and inevitably hinders the discrimination potential of learned feature embeddings.
In this work, we go beyond the intra-video correspondence learning by further considering the inter-video level embedding separation of different instance objects.
Our method is largely inspired by the recent success of contrastive learning \cite{MoCo,SimCLR}, which aims at minimizing the agreement between different augmented versions of the same image via a contrastive loss \cite{contrastiveloss}.
Nevertheless, there are two obvious gaps between contrastive learning and correspondence learning. 
First, classic contrastive learning relies on the augmented still images, but how to adapt it to the video-level correspondence scenario is rarely explored.
Second, their optimization goals are somewhat conflicting. 
Contrastive learning targets at positive concentration and negative separation, ignoring the pixel-to-pixel relevance among the positive embeddings.
In contrast, correspondence learning aims at identifying fine-grained matching.

In this work, we aim to narrow the above domain gaps by absorbing the core contrastive ideas for correspondence estimation.
To transfer the contrastive learning from the image domain to the video domain, we leverage the patch-level tracking to acquire matched image pairs in unlabeled videos.
Consequently, our method captures the real target appearance changes reside in the video sequences without augmenting the still images using empirical rules (\emph{e.g.}, scaling and rotation).
Furthermore, we propose the inter-video transformation, which is consistent with the correspondence learning in terms of the optimization goal while preserving the contrastive characteristic among different instance embeddings.
In our framework, similar to previous arts \cite{colorizition,UVC}, the image pixels should match their counterpart pixels in the current video to satisfy the self-supervision.
Besides, these pixels are also forced to mismatch the pixels in other videos to reinforce the instance-level discrimination, which is formulated in the contrastive transformation across a batch of videos, as shown in Figure~\ref{fig:1}.
%
%
%
%
By virtue of the intra-inter transformation consistency as well as the sparsity constraint for the inter-video affinity, our framework encourages the contrastive embedding learning within the correspondence framework.

In summary, the main contribution of this work lies in the contrastive framework for self-supervised correspondence learning.
1) By joint unsupervised tracking and contrastive transformation, our approach extends the classic contrastive idea to the temporal domain.
%
2) To bridge the domain gap between two diverse tasks, we propose the intra-inter transformation consistency, which differs from contrastive learning but absorbs its core motivation for correspondence tasks.
3) Last but not least, we verify the proposed approach in a series of correspondence-related tasks including video object segmentation, pose tracking, object tracking, \emph{etc}.
Our approach consistently outperforms previous state-of-the-art self-supervised approaches and is even comparable with some task-specific fully-supervised algorithms.


\section{Related Work}\label{relation work}
In this section, we briefly review the related methods including unsupervised representation learning, self-supervised correspondence learning, and contrastive learning.

{\noindent \bf Unsupervised Representation Learning}. Learning representations from unlabeled images or videos has been widely studied.
%
%
Unsupervised approaches explore the inherent information inside images or videos as the supervisory signals from different perspectives, such as frame sorting \cite{lee2017unsupervised}, image content recovering \cite{contextencoder}, deep clustering \cite{deepclustering}, affinity diffusion \cite{AffinityDiffusion}, motion modeling \cite{WatchingObjectsMove,tung2017self-supervisedMotionCapture}, and bi-directional flow estimation \cite{UnFlow}.
%
%
These methods learn an unsupervised feature extractor, which can be generalized to different tasks by further fine-tuning using a small set of labeled samples.
In this work, we focus on a sub-area in the unsupervised family, \emph{i.e.}, learning features for fine-grained pixel matching without task-specific fine-tuning. 
Our framework shares partial insight with \cite{wang2015unsupervised}, which utilizes off-the-shelf visual trackers for data pre-processing.
Differently, we jointly track and spread feature embeddings in an end-to-end manner for complementary learning. 
Our method is also motivated by the contrastive learning \cite{predictiveContrastive}, another popular framework in the unsupervised learning family.
In the following, we will detailedly discuss correspondence learning and contrastive learning.

{\noindent \bf Self-supervised Correspondence Learning}. Learning temporal correspondence is widely explored in the visual object tracking (VOT), video object segmentation (VOS), and flow estimation \cite{FlowNet} tasks.
VOT aims to locate the target box in each frame based on the initial target box, while VOS propagates the initial target mask.
To avoid expensive manual annotations, self-supervised approaches have attracted increasing attention.
%
%
%
In \cite{colorizition}, based on the frame-wise affinity, the pixel colors from the reference frame are transferred to the target frame as self-supervisory signals.      
Wang \emph{et al.} \cite{TimeCycle} conduct the forward-backward tracking in unlabeled videos and leverage the inconsistency between the start and end points to optimize the feature representation.
UDT algorithm \cite{UDT} leverages a similar bi-directional tracking idea and composes the correlation filter for unsupervised tracker training.
In \cite{TrackerSingleMovie}, an unsupervised tracker is trained via incremental learning using a single movie.
Recently, Li \emph{el al.} \cite{UVC} combine the object-level and fine-grained correspondence in a coarse-to-fine fashion and shows notable performance improvements.
In \cite{SpacetimeCorrespondence}, space-time correspondence learning is formulated as a contrastive random walk and shows impressive results.
Despite the success of the above methods, they put the main emphasis on the intra-video self-supervision.
Our approach takes a step further by simultaneously exploiting the intra-video and inter-video consistency to learn more discriminative feature embeddings.
Therefore, previous intra-video based approaches can be regarded as one part of our framework.

\begin{figure*}[t]
	\centering
	\includegraphics[width=17.7cm]{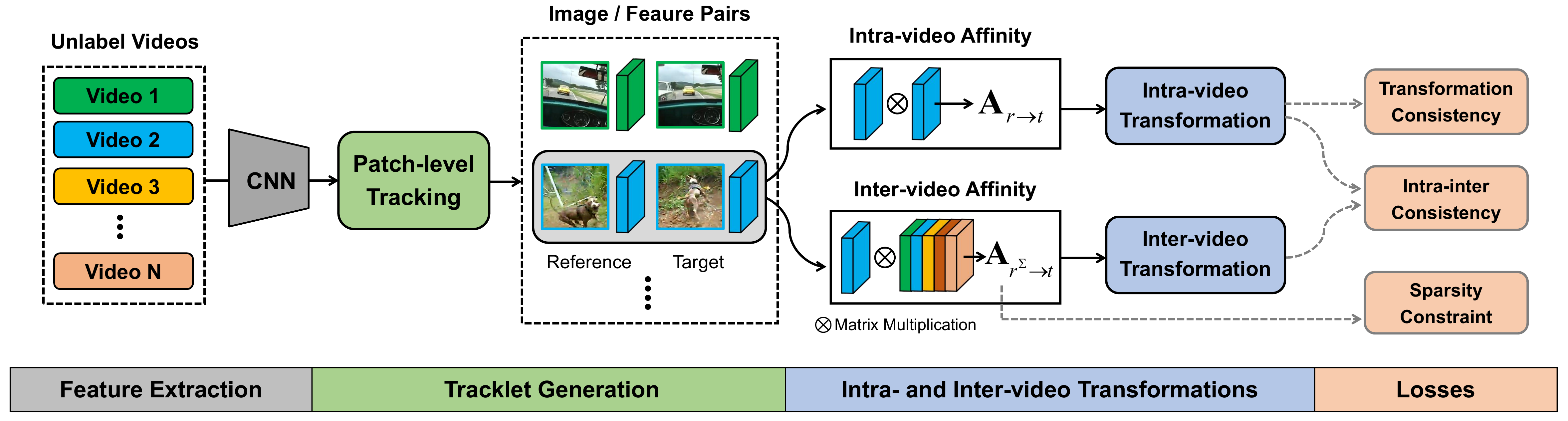}
	\caption{An overview of the proposed framework. Given a batch of videos, we first do patch-level tracking to generate image pairs. Then, intra- and inter-video transformations are conducted for each video in the mini-batch. Finally, except the intra-video self-supervision, we introduce the intra-inter consistency and sparsity constraint to reinforce the embedding discrimination.}\label{fig:2} 
\end{figure*}

{\noindent \bf Contrastive Learning}. 
Contrastive learning is a popular unsupervised learning paradigm, which aims to enlarge the embedding disagreements of different instances for representation learning \cite{predictiveContrastive,ye2019unsupervised,hjelm2019learning}. 
Based on the contrastive framework, the recent SimCLR method \cite{SimCLR} significantly narrows the performance gap between supervised and unsupervised models. 
He \emph{et al.} \cite{MoCo} propose the MoCo algorithm to fully exploit the negative samples in the memory bank.
Inspired by the recent success of contrastive learning, we also involve plentiful negative samples for discriminative feature learning.
Compared with existing contrastive methods, one major difference is our method jointly tracks and spreads feature embeddings in the video domain.
Therefore, our method captures the temporally changed appearance variations instead of manually augmenting the still images. 
Besides, instead of using a standard contrastive loss \cite{contrastiveloss}, we incorporate the contrastive idea into the correspondence task by a conceptually simple yet effective contrastive transformation mechanism to narrow the domain gap.

\section{Methodology}\label{method}

An overview of our framework is shown in Figure~\ref{fig:2}. Given a batch of videos, we first crop the adjacent image patches via patch-level tracking, which ensures the image pairs have similar contents and facilitates the later transformations. 
For each image pair, we consider the intra-video bi-directional transformation.
Furthermore, we introduce irrelevant images from other videos to conduct the inter-video transformation for contrastive embedding learning.
The final training objectives include the intra-video self-supervision, intra-inter transformation consistency, and sparsity regularization for the batch-level affinity.

\subsection{Revisiting Affinity-based Transformation}\label{}

Given a pair of video frames, the pixel colors (\emph{e.g.}, RGB values) in one frame can be copied from the pixels from another frame.
This is based on the assumption that the contents in two successive video frames are coherent.
The above frame reconstruction (pixel copy) operation can be expressed via a linear transformation with the affinity matrix $ {\bf A}_{r \to t}$, which describes the copy process from a reference frame to a target frame \cite{colorizition,liu2018switchable}. 
%

A general option for the similarity measurement in the affinity matrix is the dot product between feature embeddings. In this work, we follow previous arts \cite{colorizition,TimeCycle,UVC} to construct the following affinity matrix:
\begin{equation}\label{eq1}
	{\bf A}_{r \to t}(i,j) = \frac{\text{exp}\left({{\bf f}_t(i)}^{\top} {\bf f}_r(j)\right)}{\sum_{j} \text{exp}\left({{\bf f}_t(i)}^{\top} {\bf f}_r(j) \right) },
\end{equation}
where $ {\bf f}_t \in \mathbb{R}^{C \times N_1} $ and $ {\bf f}_r \in \mathbb{R}^{C \times N_2} $ denote flattened feature maps with $ C $ channels of target and reference frames, respectively. 
With the spatial index $ i\in[1, N_1] $ and $ j\in[1, N_2] $, ${\bf A}_{r \to t} \in \mathbb{R}^{N_1 \times N_2}$ is normalized by the softmax over the spatial dimension of $ {\bf f}_r $.

Leveraging the above affinity, we can freely transform various information from the reference frame to the target frame by $ \hat{\bf L}_t = {\bf A}_{r \to t} {\bf L}_r $, where $ {\bf L}_r $ can be any associated labels of the reference frame (\emph{e.g.}, semantic mask, pixel color, and pixel location). 
Since we naturally know the color information of the target frame, one free self-supervisory signal is color \cite{colorizition}. 
%
%
The goal of such an affinity-based transformation framework is to train a good feature extractor for affinity computation.

\subsection{Contrastive Pair Generation}\label{}
A vital step in contrastive frameworks is building positive image pairs via data augmentation. 
We free this necessity by exploring the temporal content consistency resides in the videos.
To this end, for each video, we first utilize the patch-level tracking to acquire a pair of high-quality image patches with similar content.
%
%
Based on the matched pairs, we then conduct the contrastive transformation.

Given a randomly cropped patch in the reference frame, we aim to localize the best matched patch in the target frame, as shown in Figure~\ref{fig:2}.
Similar to Eq.~\ref{eq1}, we compute a patch-to-frame affinity between the features of a random patch in the reference frame and the features of the whole target frame. 
Based on this affinity, in the target frame, we can identify some target pixels most similar to the reference pixels, and average these pixel coordinates as the tracked target center. 
We also estimate the patch scale variation following UVC approach \cite{UVC}.
%
Then we crop this patch and combine it with the reference patch to form an image pair.

\subsection{Intra- and Inter-video Transformations}\label{}

{\flushleft \bf Intra-video.} 
After obtaining a pair of matched feature maps via patch-level tracking, we compute their fined-grained affinity $ {\bf A}_{r \to t} $ according to Eq.~\ref{eq1}.
Based on this intra-video affinity, we can easily transform the image contents from the reference patch to the target patch within a single video clip.
%

{\flushleft \bf Inter-video.} 
The key success of the aforementioned affinity-based transformation lies in the embedding discrimination among plentiful subpixels to achieve the accurate label copy.
Nevertheless, within a pair of small patch regions, the image contents are highly correlated and even only cover a subregion of a large object, struggling to contain diverse visual patterns. The rarely existing negative pixels from other instance objects heavily hinder the embedding learning.
%

In the following, we improve the existing framework by introducing another inter-video transformation to achieve the contrastive embedding learning.
The inter-video affinity is defined as follows:
\begin{equation}\label{key}
	{\bf A}_{r^{\Sigma} \to t}(i,j) = \frac{\text{exp}\left({{\bf f}_t(i)}^{\top} {\bf f}_r^{\Sigma}(j)\right)}{\sum_{j} \text{exp}\left({{\bf f}_t(i)}^{\top} {\bf f}_r^{\Sigma}(j) \right) },
\end{equation}
where ${\bf f}_r^{\Sigma}$ is the concatenation of the reference features from different videos in the spatial dimension, \emph{i.e.},  $ {\bf f}_r^{\Sigma} = \text{Concat}({\bf f}_r^{1}, \cdots, {\bf f}_r^{n}) $.
For a mini-batch with $ n $ videos, the spatial index $ i\in[1, N_1] $ and $ j\in[1, nN_2] $.

\begin{figure}[t]
	\centering
	\includegraphics[width=7.8cm]{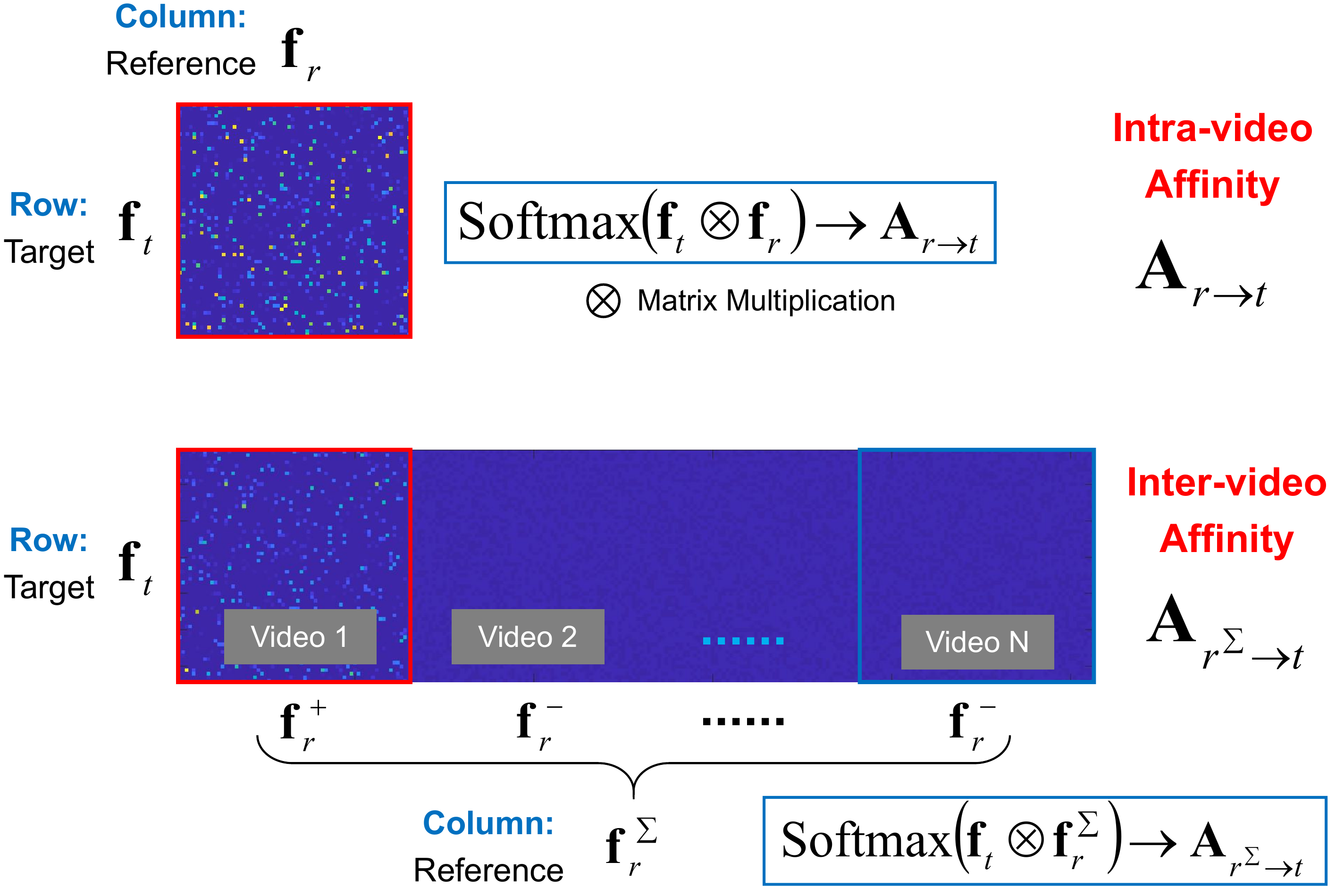}
	\caption{Comparison between intra-video affinity (top) and inter-video affinity (bottom). Best view in zoom in.}\label{fig:affinity} 
\end{figure}

{\flushleft \bf  Rationale Analysis.} Inter-video transformation is an extension of intra-video transformation. 
By decomposing the reference feature embeddings $ {\bf f}_r^{\Sigma} \in \mathbb{R}^{C \times nN_2}$ into positive and negative, $ {\bf f}_r^{\Sigma} $ can be expressed as $ {\bf f}_r^{\Sigma} = \text{Concat}({\bf f}_r^{+}, {\bf f}_r^{-}) $, where $ {\bf f}_r^{+} \in \mathbb{R}^{C \times N_2} $ denotes the only positive reference feature related to the target frame feature while $ {\bf f}_r^{-} \in \mathbb{R}^{C \times (n-1)N_2} $ is the concatenation of negative ones from unrelated videos in the mini-batch.
As a result, the computed affinity $ {\bf A}_{r^{\Sigma} \to t} \in \mathbb{R}^{N_1 \times nN_2} $ can be regarded as an ensemble of multiple sub-affinities, as shown in Figure~\ref{fig:affinity}. Our goal is to build such a batch-level affinity for discriminative representation learning.

To facilitate the later descriptions, we also divide the inter-video affinity $ {\bf A}_{r^{\Sigma} \to t}$ as a combination of positive and negative sub-affinities:
\begin{equation}\label{eq3}
	{\bf A}_{r^{\Sigma} \to t} = \text{Concat}({\bf A}_{r^{+} \to t}, {\bf A}_{r^{-} \to t}),
\end{equation}
where $ {\bf A}_{r^{+} \to t} \in \mathbb{R}^{N_1 \times N_2}$ and $ {\bf A}_{r^{-} \to t} \in \mathbb{R}^{N_1 \times (n-1) N_2} $ are the positive and negative sub-affinities, respectively.
Ideally, sub-affinity $ {\bf A}_{r^{+} \to t} $ should be close to the intra-video affinity and $ {\bf A}_{r^{-} \to t} $ is expected to be a zero-like matrix. 
Nevertheless, with the inclusion of noisy reference features $ {\bf f}_r^{-} $, the positive sub-affinity $ {\bf A}_{r^{+} \to t} $ inevitably degenerates in comparison with the intra-video affinity $ {\bf A}_{r \to t} $, as shown in Figure~\ref{fig:affinity}.
In the following, we present the intra-inter transformation consistency to encourage contrastive embedding learning within the correspondence learning task.
%

\subsection{Training Objectives}\label{training objectives}

%

To achieve the high-quality frame reconstruction, following \cite{UVC}, we pre-train an encoder and a decoder using still images on the COCO dataset \cite{COCO} to perform the feature-level transformation.
The pre-trained encoder and decoder networks are frozen without further optimization in our framework.
The goal is to train the backbone network for correspondence estimation (\emph{i.e.}, affinity computation).
In the following, the encoded features of the reference image $ {\bf I}_r $ is denoted as $ {\bf E}_r = \text{Encoder}({\bf I}_r) $.

{\flushleft \bf Intra-video Self-supervision.} Leveraging the intra-video affinity $ {\bf A}_{r \to t} $ as well as the encoded reference feature $ {\bf E}_r $, the transformed target image can be computed via $ \hat{\bf I}_{r \to t} = \text{Decoder}( {\bf A}_{r \to t} {\bf E}_r ) $.
Ideally, the transformed target frame should be consistent with the original target frame. As a consequence, the intra-video self-supervisory loss is defined as follows: 
\begin{equation}\label{key}
{\cal{L}}_{\text{self}} = \|\hat{\bf I}_{r \to t}- {\bf I}_t\|_1.
\end{equation}

\begin{figure}[t]
	\centering
	\includegraphics[width=8.5cm]{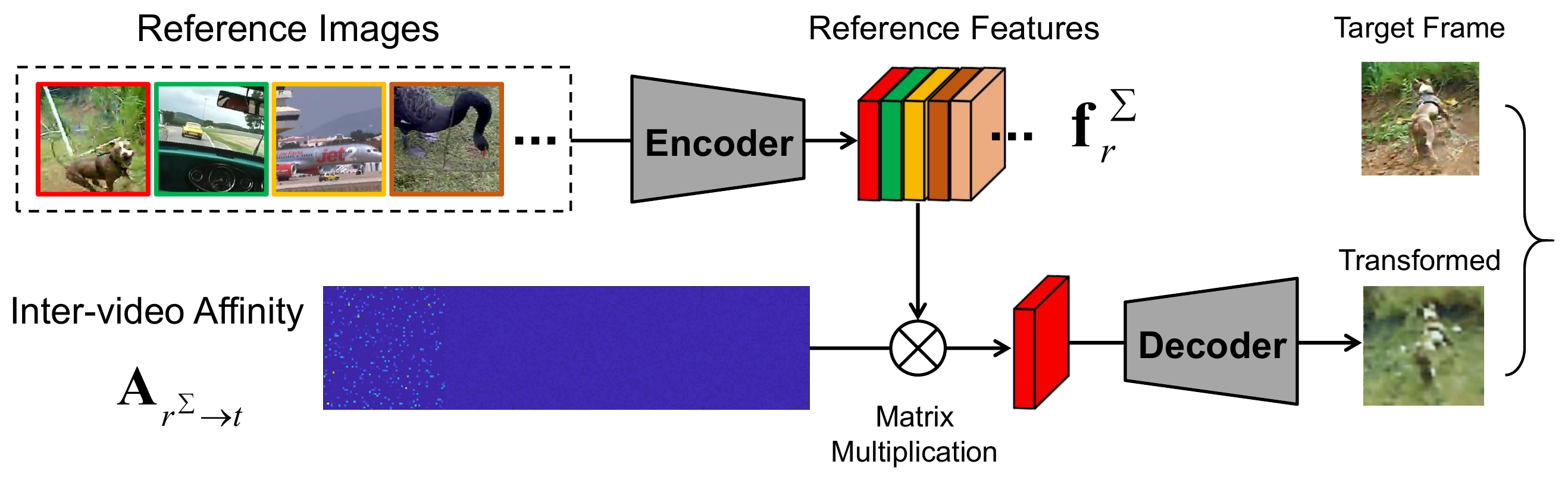}
	\caption{Illustration of the inter-video transformation.}\label{fig:transformation} 
\end{figure}

{\flushleft \bf Intra-inter Consistency.} Leveraging the inter-video affinity $ {\bf A}_{r^{\Sigma} \to t} $ and the encoded reference features $ {\bf E}_r^{\Sigma} $ from a batch of videos, \emph{i.e.}, $ {\bf E}_r^{\Sigma} = \text{Concat}({\bf E}_r^1, \cdots, {\bf E}_r^n) $, the corresponding transformed target image can be computed via $\hat{\bf I}_{r^{\Sigma} \to t} = \text{Decoder}( {\bf A}_{r^{\Sigma} \to t} {\bf E}_r^{\Sigma} ) $.
This inter-video transformation is shown in Figure~\ref{fig:transformation}.
The reference features from other videos are considered as negative embeddings.
The learned inter-video affinity is expected to exclude unrelated embeddings for transformation fidelity.
Therefore, the transformed images via intra-video affinity and inter-video affinity should be consistent: 
\begin{equation}\label{key}
{\cal{L}}_{\text{intra-inter}} = \|\hat{\bf I}_{r \to t} - \hat{\bf I}_{r^{\Sigma} \to t}\|_1.
\end{equation}
The above loss encourages both positive feature invariance and negative embedding separation.

{\flushleft \bf Sparsity Constraint.} To further enlarge the disagreements among different video features, we force the sub-affinity in the inter-video affinity $ {\bf A}_{r^{\Sigma} \to t} $ to be sparse via
\begin{equation}\label{key}
{\cal{L}}_{\text{sparse}} = \| {\bf A}_{r^- \to t}\|_1,
\end{equation}
where $ {\bf A}_{r^- \to t} $ is the negative sub-affinity in Eq.~\ref{eq3}.

{\flushleft \bf Other Regularizations.} Following previous works \cite{UVC,TimeCycle}, we also utilize the cycle-consistency (bi-directional matching) between two frames, which equals forcing the affinity matrix to be orthogonal, \emph{i.e.}, $ {\bf A}^{-1}_{r \to t} = {\bf A}_{t \to r} $.
Besides, the concentration regularization proposed in \cite{UVC} is also added.
These two regularizations are combined and denoted as $ {\cal{L}}_{\text{others}} $.

{\flushleft \bf Final Objective.} The final training objective is the combination of the above loss functions: 
\begin{equation}\label{key}
	{\cal{L}}_{\text{final}} = {\cal{L}}_{\text{self}} + {\cal{L}}_{\text{intra-inter}} + {\cal{L}}_{\text{sparse}} + {\cal{L}}_{\text{others}}.
\end{equation}
Our designed losses $ {\cal{L}}_{\text{intra-inter}} $ and $ {\cal{L}}_{\text{sparse}} $ are equally incorporated with the basic objective $ {\cal{L}}_{\text{self}} $. An overview of the training process is shown in Algorithm~1.

\subsection{Online Inference}\label{online inference}

After offline training, the pretrained backbone model is fixed during the inference stage, which is utilized to compute the affinity matrix for label transformation (\emph{e.g.}, segmentation mask). Note that the contrastive transformation is merely utilized for offline training, and the inference process is similar to the intra-video transformation.
To acquire more reliable correspondence, we further design a mutually correlated affinity to exclude noisy matching as follows:
\begin{equation}\label{key}
{\bf \widetilde{A}}_{r \to t}(i,j) = \frac{\text{exp}\left( {\bf w}(i,j) {{\bf f}_t(i)}^{\top} {\bf f}_r(j) \right)}{\sum_{j} \text{exp}\left({\bf w}(i,j) {{\bf f}_t(i)}^{\top} {\bf f}_r(j) \right) },
\end{equation}
where $ {\bf w}(i,j) \in [0,1]$ is a mutual correlation weight between two frames. Ideally, we prefer the one-to-one matching, \emph{i.e.}, one pixel in the reference frame should be highly correlated with some pixel in the target frame and vice versa.
The mutual correlation weight is formulated by: 
\begin{equation}\label{key}
{\bf w}(i,j) = \frac{{{\bf f}_t(i)}^{\top}{\bf f}_r(j)} {\max \limits_{i\in[1,N_1]} \left({{\bf f}_t(i)}^{\top} {\bf f}_r(j) \right)} \times  \frac{{{\bf f}_t(i)}^{\top}{\bf f}_r(j)} {\max \limits_{j\in[1,N_2]} \left({{\bf f}_t(i)}^{\top} {\bf f}_r(j) \right)}.
\end{equation}
The weight $ {\bf w} $ can be regarded as the affinity normalization across both reference and target spatial dimensions.
Given the above affinity between two frames, the target frame label $ \hat{\bf L}_t $ can be transformed via $ \hat{\bf L}_t = {\bf \widetilde{A}}_{r \to t} {\bf L}_r $.


\section{Experiments}\label{experiment}

We verify the effectiveness of our method on a variety of vision tasks including video object segmentation, visual object tracking, pose keypoint tracking, and human parts segmentation propagation\footnote{The source code and pretrained model will be available at \url{https://github.com/594422814/ContrastCorr}}.

\begin{algorithm}[t] \label{code1}
	\small
	\caption{Offline Training Process}
	\LinesNumbered
	\KwIn{Unlabeled video sequences.}
	\KwOut{Trained weights for the backbone network.}
	\For{each mini-batch}{
		Extract deep features of the video frames\;
		Patch-level tracking to obtain matched feature pairs\;
		
		\For{each video in the mini-batch}{
			{\tt \scriptsize {// Intra- and Inter-video transformations}}\\
			Compute intra-video affinity $ {\bf A}_{r \to t} $ (Eq.~\ref{eq1})\;
			Compute inter-video affinity $ {\bf A}_{r^{\Sigma} \to t} $ (Eq.~\ref{eq3})\;
			Conduct intra- and inter-video transformations\;
			{\tt \scriptsize {// Loss Computation}}\\
			Compute intra-video self-supervision $ {\cal L}_{\text{self}} $\;
			Compute intra-inter consistency $ {\cal L}_{\text{intra-inter}} $\;
			Compute regularization terms $ {\cal L}_{\text{sparse}} $ and $ {\cal L}_{\text{others}}$\;
		}
		Back-propagate all the losses in this mini-batch\;
	} 
\end{algorithm}

\subsection{Experimental Details} \label{sec:experiments}

{\flushleft \bf Training Details.} In our method, the patch-level tracking and frame transformations share a ResNet-18 backbone network \cite{ResNet} with the first 4 blocks for feature extraction. 
The training dataset is TrackingNet \cite{2018trackingnet} with about 30k video. Note that previous works \cite{TimeCycle, UVC} use the Kinetics dataset \cite{Kinetics}, which is much larger in scale than TrackingNet.
%
%
Our framework randomly crops and tracks the patches of 256$\times$256 pixels (\emph{i.e.}, patch-level tracking), and further yields a 32$\times$32 intra-video affinity (\emph{i.e.}, the network stride is 8).
The batch size is 16. Therefore, each positive embedding contrasts with 15$\times$(32$\times$32$ \times $2) = 30720 negative embeddings. 
Since our method considers pixel-level features, a small batch size also involves abundant contrastive samples.
We first train the intra-video transformation (warm-up stage) for the first 100 epochs and then train the whole framework in an end-to-end manner for another 100 epochs.
The learning rate of both two stages is $ 1\times10^{-4} $ and will be reduced by half every 40 epochs.
The training stage takes about one day on 4 Nvidia 1080Ti GPUs.

{\noindent \bf Inference Details.} For a fair comparison, we use the same testing protocols as previous works \cite{TimeCycle, UVC} in all tasks. 

\subsection{Framework Effectiveness Study} \label{sec:ablation}

In Table~\ref{table:ablation study}, we show ablative experiments of our method on the DAVIS-2017 validation dataset \cite{DAVIS2017}.
The evaluation metrics are Jacaard index $ \cal{J} $ and contour-based accuracy $ \cal{F} $.
As shown in Table~\ref{table:ablation study}, without the intra-video guidance, inter-video transformation alone for self-supervision yields unsatisfactory results due to overwhelming noisy/negative samples.
With only intra-video transformation, our framework is similar to the previous approach \cite{UVC}.
By jointly employing both of these two transformations under an intra-inter consistency constraint, our method obtains obvious performance improvements of 3.2\% in $ \cal{J} $ and  3.4\% in $\cal{F} $.
The sparsity term of inter-video affinity encourages the embedding separation and further improves the results.

In Figure \ref{fig:comparison}, we further visualize the comparison results of our method with and without contrastive transformation.
As shown in the last row of Figure \ref{fig:comparison}, only intra-video self-supervision fails to effectively handle the challenging scenarios with distracting objects and partial occlusion.
By involving the contrastive transformation, the learned feature embeddings exhibit superior discrimination capability for instance-level separation.

\makeatletter
\def\hlinew#1{%
	\noalign{\ifnum0=`}\fi\hrule \@height #1 \futurelet
	\reserved@a\@xhline}
\makeatother

\setlength{\tabcolsep}{2pt}
\begin{table}[t]
	\scriptsize
	\begin{center}	
		\begin{tabular*}{8.0 cm} {@{\extracolsep{\fill}}lcccc|cc}
			&Intra-video &Inter-video &Sparsity & Mutual & $ \cal{J} $(Mean) &$ \cal{F} $(Mean)  \\
			&Transformation &Transformation &Constraint  & Correlation & &   \\
            &$ {\cal L}_{\text{self}} + {\cal L}_{\text{others}} $ &$ {\cal L}_{\text{intra-inter}} $ &$ {\cal L}				_{\text{sparse}} $ & \\
			\hlinew{1pt}
			&$\checkmark$ & & & &55.8 &60.3 \\
			&$\checkmark$ &$\checkmark$  & &&59.0 &63.7\\
			&$\checkmark$ &$\checkmark$  &$\checkmark$ &&59.2 &64.0 \\
			&$\checkmark$ &$\checkmark$  &$\checkmark$ &$\checkmark$& {\bf 60.5} &{\bf 65.5} \\
		\end{tabular*}
		\caption{Analysis of each component of our method on the DAVIS-2017 validation dataset.} \label{table:ablation study}
	\end{center}
\end{table}

\begin{figure}[t]
	\centering
	\includegraphics[width=8.3cm]{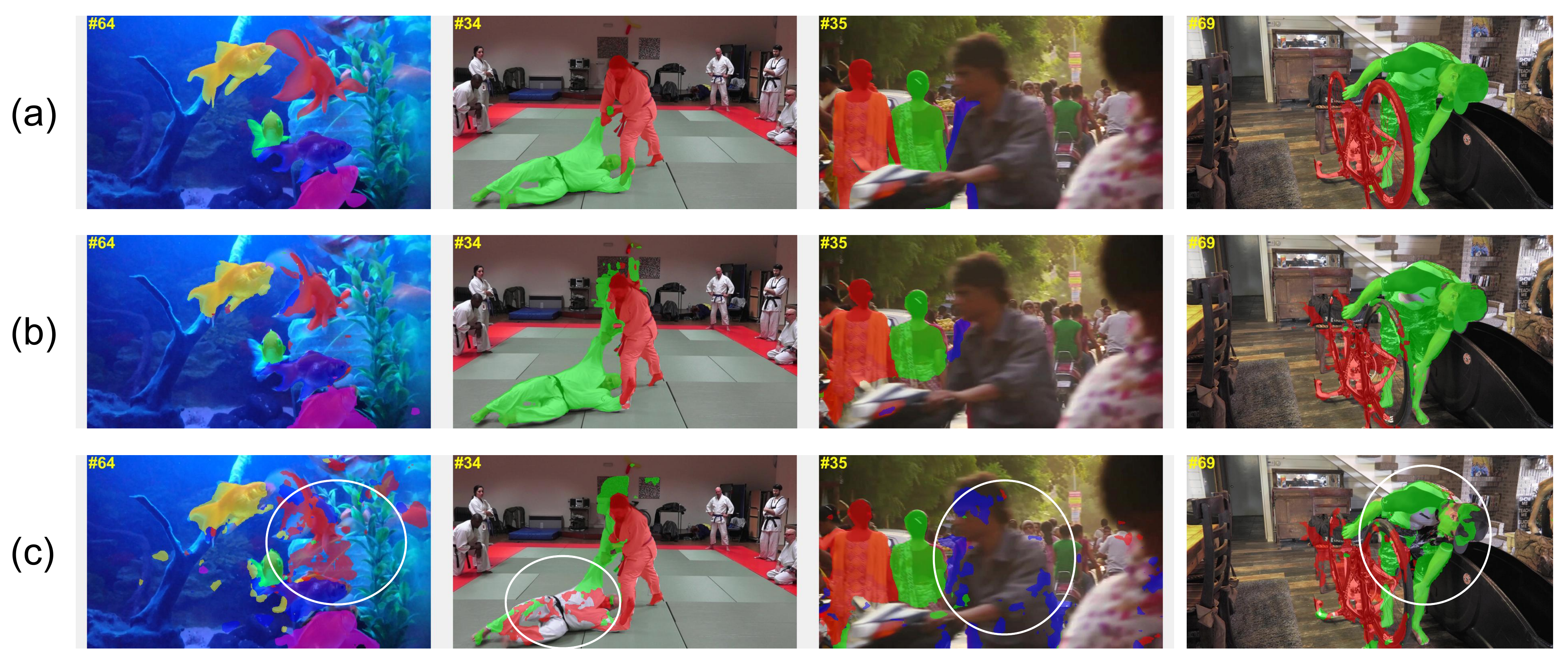}
	\caption{(a) Ground-truth results. (b) Results of the model with both intra- and inter-video transformations. (c) Results of the model without inter-video contrastive transformation, where the failures are highlighted by white circles. 
	}\label{fig:comparison} 
\end{figure}

\begin{figure*}[t]
	\centering
	\includegraphics[width=17.8cm]{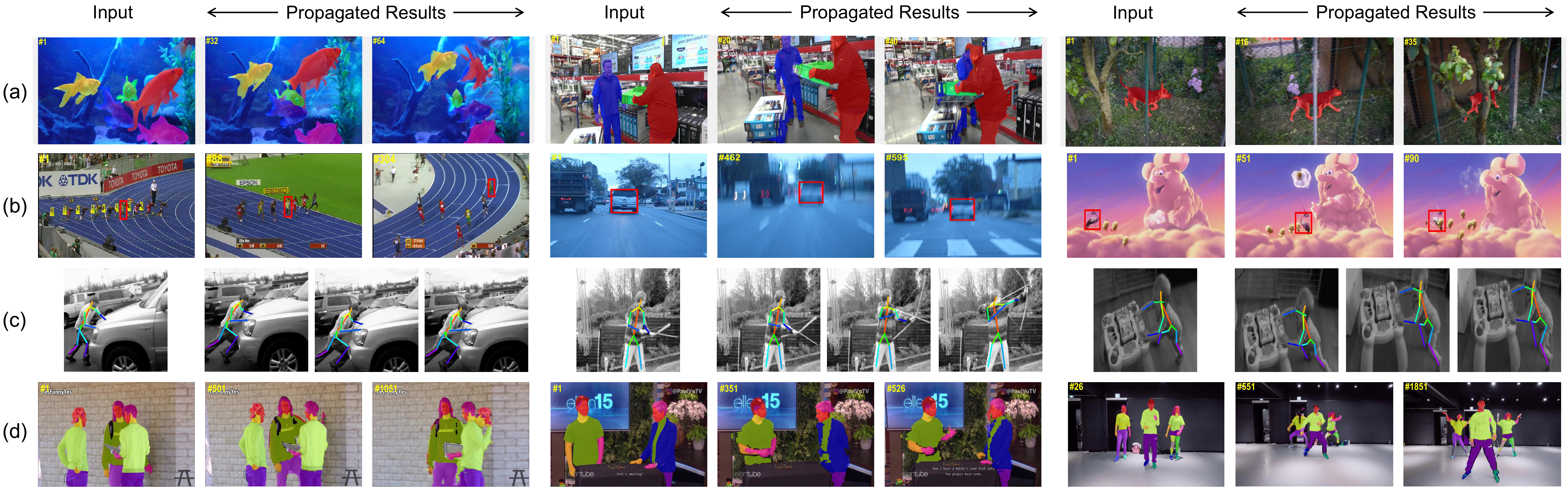}
	\caption{Experimental results of our method. (a) Video object segmentation on the DAVIS-2017. (b) Visual object tracking on the OTB-2015. (c) Pose keypoint tracking on the J-HMDB. (d) Parts segmentation propagation on the VIP.}\label{fig:examples} 
\end{figure*}

\setlength{\tabcolsep}{2pt}
\begin{table}[t]
	\scriptsize
	\begin{center}
		\begin{tabular*}{8.0 cm} {@{\extracolsep{\fill}}l|c|cc}
			Model & Supervised & $ \cal{J} $(Mean) &$ \cal{F} $(Mean) \\
			\hlinew{1pt}
			Transitive Inv. \cite{Transitive}  &  &32.0 &26.8 \\
			DeepCluster \cite{deepclustering} &  &37.5 &33.2 \\
			Video Colorization \cite{colorizition} & &34.6 &32.7 \\
			Time-Cycle \cite{TimeCycle}  &  &41.9 &39.4 \\
			CorrFlow \cite{CorrFlow} & &48.4 &52.2 \\
			UVC (480p) \cite{UVC}  &  &56.3 &59.2 \\
			UVC (560p) \cite{UVC}  &  &56.7 &60.7 \\
			MAST \cite{MAST} & &{\bf 63.3} &{\bf 67.6} \\
			{\bf ContrastCorr (Ours)}  & &60.5 &65.5 \\
			\hline
			ResNet-18 \cite{ResNet} &$\checkmark$ &49.4 &55.1  \\
			OSVOS \cite{OSVOS}  &$\checkmark$ &56.6 &63.9  \\
			FEEVOS \cite{voigtlaender2019feelvos:} &$\checkmark$ &69.1 &74.0 \\
		\end{tabular*}
		\caption{Evaluation on video object segmentation on the DAVIS-2017 validation dataset. The evaluation metrics are region similarity $ \cal{J} $ and contour-based accuracy $ \cal{F} $ .} \label{table:DAVIS2017}	
	\end{center}
\end{table}

\setlength{\tabcolsep}{2pt}
\begin{table}[t]
	\scriptsize
	\begin{center}
		\begin{tabular*}{8.0 cm} {@{\extracolsep{\fill}}l|c|cc}
			Model & Supervised & DP@20pixel & AUC \\
			\hlinew{1pt}
			KCF (HOG feature) \cite{KCF} & &69.6 &48.5 \\
			UL-DCFNet \cite{TrackerSingleMovie} & &75.5 &58.4 \\
			UDT \cite{UDT} & &76.0 &59.4 \\
			UVC \cite{UVC}  &  &- &59.2 \\
			LUDT \cite{LUDT} & &76.9 &60.2 \\
			{\bf ContrastCorr (Ours)}  & &{\bf 77.2} &{\bf 61.1} \\
			\hline
			ResNet-18 + DCF \cite{ResNet} &$\checkmark$ &49.4 &55.6  \\
			SiamFC \cite{SiamFc} &$\checkmark$ &77.1 &58.2  \\
			DiMP-18 \cite{DiMP}  &$\checkmark$ &87.1  &66.2  \\
		\end{tabular*}
		\caption{Evaluation on video object tracking on the OTB-2015 dataset. The evaluation metrics are distance precision (DP) and area-under-curve (AUC) score of the success plot.} \label{table:OTB2015}
	\end{center}
\end{table}

\begin{table}[t]
	\scriptsize
	\begin{center}
		\begin{tabular*}{7.8 cm} {@{\extracolsep{\fill}}l|c|cc}
			Model & Supervised & PCK@.1 & PCK@.2 \\
			\hlinew{1pt}
			SIFT Flow \cite{SIFTFlow} &  &49.0 &68.6 \\
			Transitive Inv. \cite{Transitive}  & &43.9 &67.0 \\
			DeepCluster \cite{deepclustering} & &43.2 &66.9 \\
			Video Colorization \cite{colorizition}  & &45.2 &69.6 \\
			Time-Cycle \cite{TimeCycle}  & &57.3 &78.1 \\
			CorrFlow \cite{CorrFlow} & &58.5 &78.8 \\
			UVC \cite{UVC}  & &58.6 &79.8 \\
			{\bf ContrastCorr (Ours)} & &{\bf 61.1} &{\bf 80.8} \\
			\hline
			ResNet-18 \cite{ResNet} &$\checkmark$ &53.8 &74.6  \\
			Thin-Slicing Network \cite{thin-slicing} &$\checkmark$ &68.7 &92.1  \\
		\end{tabular*}
		\caption{Keypoints propagation on J-HMDB. The evaluation metric is PCK at different thresholds.} \label{table:JHMDB}	
	\end{center}
\end{table}

\setlength{\tabcolsep}{2pt}
\begin{table}[t]
	\scriptsize
	\begin{center}	
		\begin{tabular*}{7.6 cm} {@{\extracolsep{\fill}}l|c|cc}
			Model &Supervised &mIoU & $ \text{AP}^{r}_{\text{vol}} $~ \\
			\hlinew{1pt}
			SIFT Flow \cite{SIFTFlow} &  &21.3 &10.5 \\
			Transitive Inv. \cite{Transitive}  &  &19.4 &5.0 \\
			DeepCluster \cite{deepclustering}  &  &21.8 &8.1 \\
			Time-Cycle \cite{TimeCycle}  &  &28.9 &15.6 \\
			UVC \cite{UVC}  &  &34.1 &17.7 \\
			{\bf ContrastCorr (Ours)}  & &{\bf 37.4} &{\bf 21.6} \\
			\hline
			ResNet-18 \cite{ResNet}  &$\checkmark$ &31.8 &12.6  \\
			FGFA \cite{zhu2017flow-guided} &$\checkmark$ &37.5 &23.0  \\
			ATEN \cite{VIP} &$\checkmark$ &37.9 &24.1  \\
		\end{tabular*}
		\caption{Evaluation on propagating human part labels in Video Instance-level Parsing (VIP) dataset. The evaluation metrics are semantic propagation with mIoU and part instance propagation in $ \text{AP}^{r}_{\text{vol}} $.} \label{table:VIP}
	\end{center}
\end{table}

\subsection{Comparison with State-of-the-art Methods} 

%

{\noindent \bf Video Object Segmentation on the DAVIS-2017.} 
DAVIS \cite{DAVIS2017} is a video object segmentation (VOS) benchmark.
We evaluate our method on the DAVIS-2017 validation set following Jacaard index $ \cal{J} $ (IoU) and contour-based accuracy $ \cal{F} $.
Table~\ref{table:DAVIS2017} lists quantitative results. 
Our model performs favorably against the state-of-the-art self-supervised methods including Time-Cycle \cite{TimeCycle}, CorrFlow \cite{CorrFlow}, and UVC \cite{UVC}. 
Specifically, with the same experimental settings (\emph{e.g.}, frame input size and recurrent reference strategy), our model surpasses the recent top-performing UVC approach by 3.8\% in $ \cal{J} $ and 4.8\% in $ \cal{F} $. 
The recent MAST approach \cite{MAST} obtains impressive results by leveraging a memory mechanism, which can be added to our framework for further performance improvement.
From Figure~\ref{fig:examples} (first row), we can observe that our method is robust in handling distracting objects and partial occlusion.

Compared with the fully-supervised ResNet-18 network trained on ImageNet with classification labels, our method exhibits much better performance.
It is also worth noting that our method even surpasses the recent fully-supervised methods such as OSVOS. 

{\noindent \bf Video Object Tracking on the OTB-2015.} 
OTB-2015 \cite{OTB-2015} is a visual tracking benchmark with 100 challenging videos.
We evaluate our method on OTB-2015 under distance precision (DP) and area-under-curve (AUC) metrics.
Our model learns robust feature representations for fine-grained matching, which can be combined with the correlation filter \cite{KCF,DSST} for robust tracking.
Without online fine-tuning, we integrate our model into a classic tracking framework based on the correlation filter, \emph{i.e.,} DCFNet \cite{DCFNet}.
The comparison results are shown in Table~\ref{table:OTB2015}.
Note that UDT \cite{UDT} is the recently proposed unsupervised tracker trained with the correlation filter in an end-to-end manner.
Without end-to-end optimization, our model is still robust enough to achieve superior performance in comparison with UDT.
Our method also outperforms the classic fully-supervised trackers such as SiamFC.
As shown in Figure~\ref{fig:examples} (second row), our model can well handle the motion blur, deformation, and similar distractors.

{\noindent \bf Pose Keypoint Propagation on the J-HMDB.} 
We evaluate our model on the pose keypoint propagation task on the validation set of J-HMDB \cite{JHMDB}. 
Pose keypoint tracking requires precise fine-grained matching, which is more challenging than the box-level or mask-level propagation in the VOT/VOS tasks.
Given the initial frame with 15 annotated human keypoints, we propagate them in the successive frames.
The evaluate metric is the probability of correct keypoint (PCK), which measures the percentage of keypoints close to the ground-truth in different thresholds.
We show comparison results against the state-of-the-art methods in Table~\ref{table:JHMDB} and qualitative results in Figure~\ref{fig:examples} (third row).
Our method outperforms all previous self-supervised methods such as Time-Cycle, CorrFlow, and UVC (Table~\ref{table:JHMDB}).
Furthermore, our approach significantly outperforms pre-trained ResNet-18 with ImageNet supervision.

{\noindent \bf Semantic and Instance Propagation on the VIP.} 
Finally, we evaluate our method on the Video Instance-level Parsing
(VIP) dataset \cite{VIP}, which includes dense human parts segmentation masks on both the semantic and instance levels. 
%
%
We conduct two tasks in this benchmark: semantic propagation and human part propagation with instance identity.
For the semantic mask propagation, we propagate the semantic segmentation maps of human parts (\emph{e.g.}, heads, arms, and legs) and evaluate performance via the mean IoU metric. 
For the part instance propagation task, we propagate the instance-level segmentation of
human parts (\emph{e.g.}, different arms of different persons) and evaluate performance via the instance-level human parsing metric: mean Average Precision (AP).
Table~\ref{table:VIP} shows that our method performs favorably against previous self-supervised methods.
For example, our approach outperforms the previous best self-supervised method UVC by 3.3\% mIoU in semantic propagation and 3.9\% in human part propagation.
Besides, our model notably surpasses the ResNet-18 model trained on ImageNet with classification labels.
Finally, our method is comparable with the fully-supervised ATEN algorithm \cite{VIP} designed for this dataset.

\section{Conclusion} \label{conclusion}
In this work, we focus on the correspondence learning using unlabeled videos.
Based on the well-studied intra-video self-supervision, we go one step further by introducing the inter-video transformation to achieve contrastive embedding learning.
The proposed contrastive transformation encourages embedding discrimination while preserving the fine-grained matching characteristic among positive embeddings.
Without task-specific fine-tuning, our unsupervised model shows satisfactory generalization on a variety of temporal correspondence tasks.
Our approach consistently outperforms previous self-supervised methods and is even comparable with the recent fully-supervised algorithms.

{ {\flushleft \bf Acknowledgements.} The work of Wengang Zhou was supported in part by the National Natural Science Foundation of China under Contract 61822208, Contract U20A20183, and Contract 61632019; and in part by the Youth Innovation Promotion Association CAS under Grant 2018497. The work of Houqiang Li was supported by NSFC under Contract 61836011.}

%
%
%
%

{
	\bibliographystyle{aaai}
	\bibliography{reference}
}

\newpage

\appendix
\renewcommand{\appendixname}{Appendix~\Alph{section}}

\section{Inference Details}

In the inference stage, we leverage the computed affinity matrix to transform different types of inputs, \emph{e.g.}, segmentation masks and pose keypoints. 
Similar to Time-Cycle and UVC, we adopt the same recurrent inference strategy to propagate the ground-truth result from the first frame,
as well as the predicted results from the preceding $ L $ frames onto the target frame. We average all
$ L+1 $ predictions to obtain the final propagated map.
Following previous works, $ L $ is set to 1 for the VIP dataset and 7 for all the rest benchmarks.
For fair comparisons, following Time-Cycle and UVC, we also use the k-NN propagation schema and set k = 5 for all tasks.
More details can be found in the source code.

\section{Transformation Results}

\begin{figure*}
	\centering
	\includegraphics[width=15.5cm]{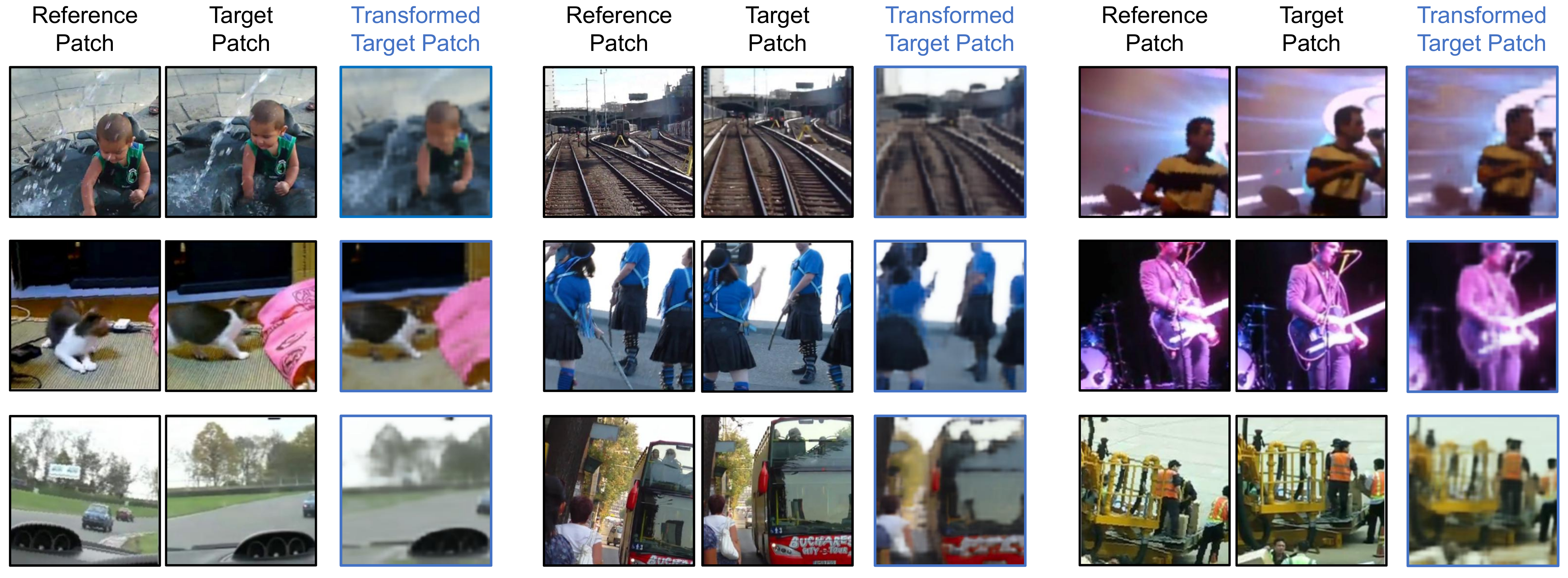}
	\caption{Examples of our tracked image pairs and transformed patches.}\label{fig:img_pair}
	\vspace{-0.0in}
\end{figure*}

In Figure~\ref{fig:img_pair}, we exhibit some examples of our tracked image pairs.
In our framework, we first randomly crop a reference patch in the reference frame and then conduct the patch-level tracking to form a pair of matched images.
As shown in Figure~\ref{fig:img_pair}, the image pairs have similar contents, which facilitate further intra- and inter-video transformations.
Thanks to the patch-level tracking, our image pairs contain the real target appearance changes (\emph{e.g.}, person view/pose changes), which differs from conventional contrastive methods based on the manually designed rules (\emph{e.g.}, flip and rotation) to form image pairs.

In Figure~\ref{fig:img_pair}, we also show the inter-video transformation results of our approach.
The transformed images yield almost identical contents in comparison with the target patch, which affirms that our affinity matrix achieves reliable correspondence matching.

\begin{figure*}
	\centering
	\includegraphics[width=16cm]{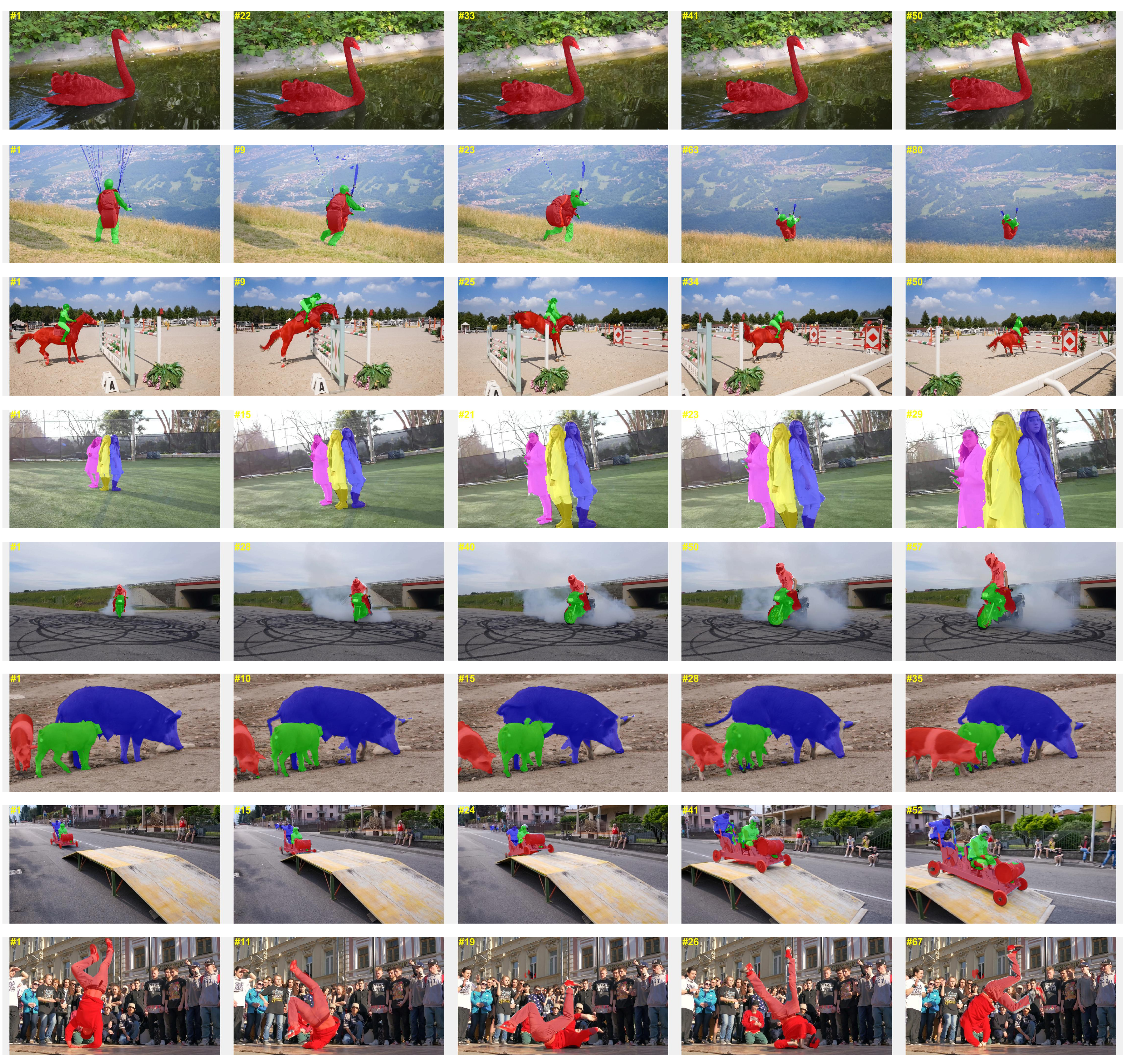}
	\caption{More results on the DAVIS-2017 validation dataset.}\label{fig:examples}
	\vspace{-0.0in}
\end{figure*}

\begin{figure*}
	\centering
	\includegraphics[width=17.5cm]{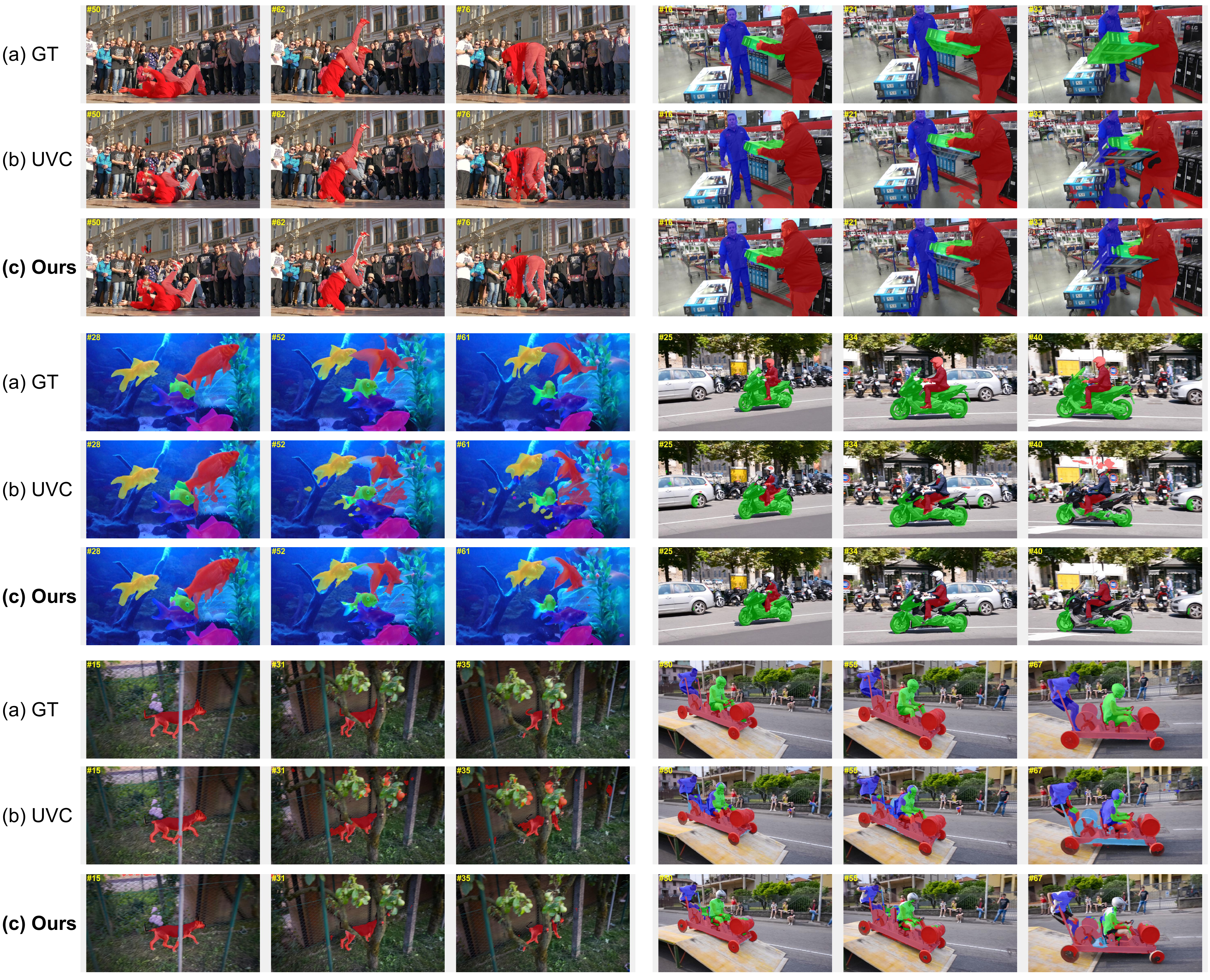}
	\caption{(a) Ground-truth segmentation results. (b) Results of UVC, which represents the current state-of-the-art performance of self-supervised correspondence methods. (c) Our results. By virtue of contrastive transformation, our approach shows superior results in comparison with previous intra-video based methods.}\label{fig:comparison}
	\vspace{-0.0in}
\end{figure*}

\section{Additional VOS Results}

In Figure~\ref{fig:examples}, we show more results of our approach on the DAVIS-2017 validation dataset.
From Figure~\ref{fig:examples}, we can observe that our method is able to accurately propagate the segmentation masks in challenging scenarios.

UVC algorithm represents the current state-of-the-art self-supervised correspondence approach based on the intra-video transformation paradigm.
In contrast, our method further exploits the inter-video level transformation to reinforce instance-level embedding discrimination.
In Figure~\ref{fig:comparison}, we further compare our approach with UVC. 
As shown in Figure~\ref{fig:comparison}, compared with UVC, our approach better handles the challenging scenarios such as occlusion, deformation, and similar distractors.

\end{document}